\documentclass[letterpaper, 10 pt, conference]{ieeeconf}
\IEEEoverridecommandlockouts
\title{\LARGE \bf
Distributionally Robust Policy Learning via \\ Adversarial Environment Generation}

\author{Allen Z. Ren$^{1}$ and Anirudha Majumdar$^{1}$% <-this % stops a space
\thanks{*This work was supported by the Toyota Research Institute and the Office of Naval Research}% <-this % stops a space
\thanks{$^{1}$Allen Z. Ren and Anirudha Majumdar are with Mechanical and Aerospace Engineering Department, Princeton University, United States {\tt\small \{allen.ren, ani.majumdar\}@princeton.edu}}%
}

\let\labelindent\relax  % error Command \labelindent already defined from IEEE
\usepackage{graphicx, float, url, wrapfig}
\usepackage{amsmath, amssymb, bbm, mathtools}
\usepackage[font=footnotesize]{caption}
\usepackage{makecell, tabularx, multirow,  adjustbox}

\newcommand{\bftab}{\fontseries{b}\selectfont}
\usepackage{scalerel}
\usepackage{bibunits}
\usepackage{booktabs}
\usepackage{xcolor}

\usepackage{algorithm}
\usepackage{algpseudocode}
\algnewcommand{\Comment}[1]{\hfill// \eqparbox{COMMENT\thealgorithm}{#1}}
\usepackage{enumitem}

\DeclareMathOperator*{\minimize}{minimize}

\newcommand{\newsup}{\mathop{\smash{\mathrm{sup}}}}
\newcommand{\newarginf}{\mathop{\smash{\mathrm{arginf}}}}

\newtheorem{definition}{\bf Definition}

\begin{document}

\maketitle
\thispagestyle{empty}
\pagestyle{empty}

%%%%%%%%%%%%%%%%%%%%%%%%%%%%%%%%%%%%%%%%%%%%%%%%%%%%%%%%%%%%%%%%%%%%%%%%%%%%%%%%
\begin{abstract}

Our goal is to train control policies that generalize well to unseen environments. Inspired by the Distributionally Robust Optimization (DRO) framework, we propose DRAGEN --- Distributionally Robust policy learning via Adversarial Generation of ENvironments --- for iteratively improving robustness of policies to realistic distribution shifts by generating adversarial environments.
The key idea is to learn a generative model for environments whose latent variables capture cost-predictive and realistic variations in environments.
We perform DRO with respect to a Wasserstein ball around the empirical distribution of environments by generating realistic adversarial environments via gradient ascent on the latent space.
We demonstrate strong Out-of-Distribution (OoD) generalization in simulation for (i) swinging up a pendulum with onboard vision and (ii) grasping realistic 3D objects. Grasping experiments on hardware demonstrate better sim2real performance compared to domain randomization.
% 2D/
\end{abstract}


%%%%%%%%%%%%%%%%%%%%%%%%%%%%%%%%%%%%%%%%%%%%%%%%%%%%%%%%%%%%%%%%%%%%%%%%%%%%%%%%
\begin{bibunit}[IEEEtran.bst]
% \vspace{-5pt}
\section{Introduction}
\label{sec:intro}
% \vspace{-5pt}

\begin{figure*}[t]
\centering
\includegraphics[width=1.0\textwidth]{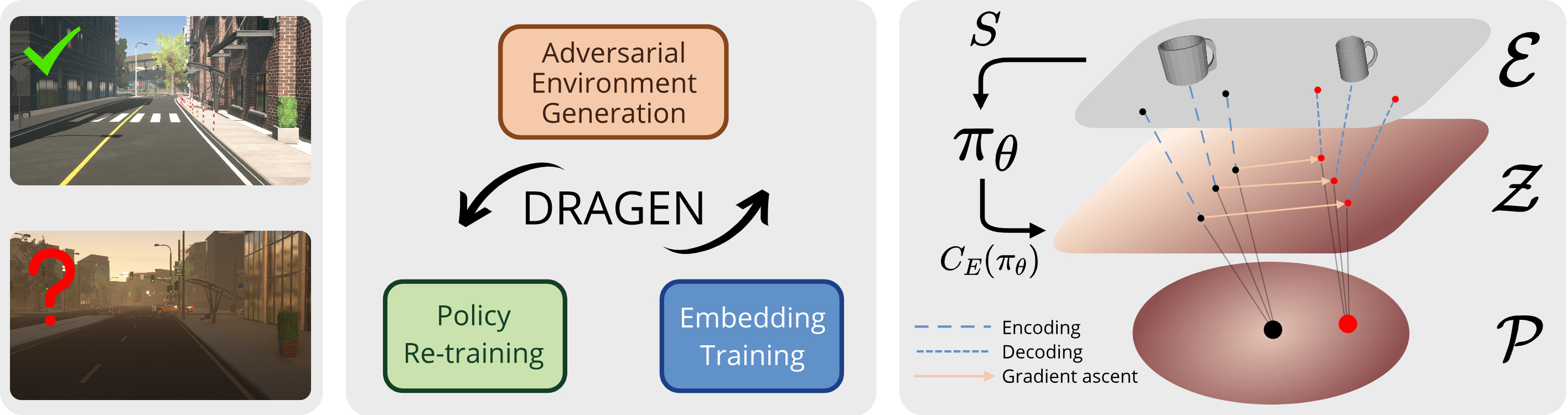}
\vspace{-12pt}
\caption{\footnotesize (Left) Control policies often fail under distributional shift of environments such as changing lighting conditions. (Middle) Our proposed framework, DRAGEN, trains policies to generalize to such test environments (Sec.~\ref{sec:experiments}). It alternates between training a cost-predictive latent space of a generative model, generating adversarial environments via the latent space, and re-training the policy using the augmented dataset. (Right) The training dataset $S$ consists of environments $E$ (e.g., mugs to be grasped) in $\mathcal{E}$. They are embedded in the latent space $\mathcal{Z}$ of a generative model. We consider the resulting set of latent embeddings as a discrete distribution around which the uncertainty set $\mathcal{P}$ is defined. We apply recent progress in DRO \cite{sinha2017certifying} for performing worst-case optimization over $\mathcal{P}$ by perturbing the support of the discrete distribution.
Costs $C_E(\pi_\theta)$ incurred by the policy trains $\mathcal{Z}$ to be cost-predictive, and allows for gradient ascent on $\mathcal{Z}$. Decoding adversarial latent variables generates realistic adversarial environments, such as mugs with smaller openings.}
\label{fig:anchor}
\vspace{-15pt}
\end{figure*}

One of the fundamental challenges for learning-based control of robots is the severely limited ability of trained policies to generalize beyond the specific distribution of environments they were trained on. For example, imagine a home-robot with manipulation capabilities that has been trained on a dataset containing thousands of objects. How likely is this system to succeed when deployed in different homes containing objects that the system has never encountered before? Similarly, how likely is a vision-based navigation policy for a drone or autonomous vehicle to succeed when deployed in environments with varying weather conditions, lighting (Fig.~\ref{fig:anchor}), or obstacle densities? Unfortunately, current techniques for learning-based control of robots (e.g., those based on deep reinforcement learning) can fail dramatically when faced with even mild distribution shifts \cite{ sunderhauf2018limits}. 

In this work, we pose the problem of learning policies with \emph{Out-of-Distribution} (OoD) generalization capabilities in the framework of \emph{Distributionally Robust Optimization (DRO)}; given a dataset of environments (e.g., objects in the case of manipulation), our goal is to learn a policy that minimizes the \emph{worst-case expected cost}
across a set $\mathcal{P}$ of distributions around the empirical distribution:
\begin{equation}
    \underset{\theta \in \Theta}{\inf} \ \newsup_{P \in \mathcal{P}} \ \underset{E \sim P}{\mathbb{E}} [C_E(\pi_\theta)],
\label{eq:minimax}
\end{equation}
where $\theta$ are the parameters of the policy (e.g., weights of a neural network), and $C_E(\pi_\theta)$ is the cost incurred by policy $\pi_\theta$ when deployed in environment $E$ (see Sec.~\ref{sec:formulation} for a formal problem formulation). 
One of the key ingredients of this formalism is the choice of the set $\mathcal{P}$ of distributions over which one performs worst-case optimization. This choice is crucial in robotics applications and must satisfy two important criteria. First, the set $\mathcal{P}$ should contain \emph{realistic} distributions. Second, it should be \emph{broad enough} to encompass distributions that vary in \emph{task-relevant} features (e.g., geometric features of the objects in grasping task are task-relevant, while colors are not) and thus help improve generalization to real-world OoD environments. The main technical insight of our work is to combine ideas from the theory of \emph{Wasserstein metrics} with advances in generative modeling and adversarial training to propose a DRO framework for learning policies that are robust to realistic distribution shifts. We highlight key features of our approach next.

\textbf{Statement of Contributions.} The primary contribution of this work is to propose DRAGEN, a framework based on DRO for iteratively improving the robustness of policies to realistic distribution shifts by generating adversarial environments (Fig.~\ref{fig:anchor}). To this end, we make four specific contributions.
\vspace{-5pt}
\begin{itemize}[leftmargin=*]
    \item Develop an approach for learning a generative model for environments (using a given training dataset) whose latent variables capture task-relevant and realistic variations in environments (Sec.~\ref{sec:realistic}, \ref{sec:task-relevant}). This is achieved by training the latent variables to be \emph{cost-predictive}
%  of the generative model 
    % (in terms of costs incurred by the current policy) 
    and regularizing the Lipschitz constant of the cost predictor; this ensures that distances in the latent space correspond to task-relevant differences in environments. % (as measured by differences in the cost incurred by the current policy).
    \item Propose a method for specifying the set $\mathcal{P}$ of distributions over which we perform DRO as a \emph{Wasserstein ball} defined with respect to distance in the latent space (Sec.~\ref{sec:realistic}). This ensures that $\mathcal{P}$ contains distributions over task-relevant and realistic variations in environments. We also provide a distributionally robust bound on the worst-case expected predicted cost of distributions in $\mathcal{P}$.
    \item Develop an algorithm for performing DRO with respect to the Wasserstein ball by adversarial generation of environments (Sec.~\ref{sec:optimization}). Our overall approach then iteratively improves the policy by alternating between (i) re-training the generative model, (ii) generating adversarial environments for DRO, and (iii) re-training the policy using the augmented dataset.
    \item Demonstrate the ability of our approach to learn policies with strong OoD generalization in simulation for (i) swinging up a pendulum with onboard vision and (ii) grasping realistic 2D/3D objects (Sec.~\ref{sec:experiments}). We also validate our approach on grasping experiments in hardware and demonstrate that DRAGEN outperforms domain randomization in sim2real transfer. 
\end{itemize}

\vspace{-10pt}
\subsection{Related work}
% \vspace{-7pt}
\textbf{Distributionally robust optimization.} Our work is inspired by the DRO framework in supervised learning \cite{sinha2017certifying, blanchet2019quantifying, esfahani2018data, namkoong2016stochastic}, which minimizes the risk under worst-case distributional shift of data (similar to \eqref{eq:minimax}).
Recent progress provides a direct solution to the Lagrangian relaxation of the formulation \cite{sinha2017certifying}, which suits our approach (Sec.~\ref{sec:approach}). In terms of the choice of uncertainty set $\mathcal{P}$, \cite{sinha2017certifying, esfahani2018data} defines it using Wasserstein distance; this allows $\mathcal{P}$ to include distributions with different support and can thus provide robustness to unseen data. In contrast to \cite{volpi2018generalizing} where the Wasserstein distance is defined on the semantic space (output of the last hidden layer) of the classifier, we define Wasserstein distance based on distance on the latent space of a generative model; this allows us to capture realistic distributional shifts of environments. In addition, we train a Lipschitz-regularized cost predictor from the low-dimensional latent space. This provides structure to the latent space by ensuring that nearby points correspond to environments with similar costs, and improves distributional robustness.

% that unlike $f$-divergences \cite{namkoong2016stochastic}, 
% \vspace{-3pt}
\textbf{Environment augmentation in policy learning.} Domain Randomization techniques generate new training environments by randomizing pre-specified parameters such as object textures or lighting intensities \cite{tobin2017domain, mehta2020active}, or randomly chaining shape primitives into new objects \cite{tobin2018domain} for grasping. Similarly, data augmentation techniques such as random cutout and cropping \cite{kostrikov2020image, laskin2020reinforcement} have been applied to vision-based reinforcement learning (RL). Despite their simplicity, both types of techniques can be inefficient and do not necessarily generate realistic environments. For instance, training on randomly generated objects leads to worse performance in grasping realistic objects than training on the same number of realistic objects \cite{tobin2018domain}. Another line of work generates increasingly difficult and complex environments \cite{dennis2020emergent, wang2019paired} using minimax formulations based on the agent's performance with the current policy, which are similar to our approach but do not focus on OoD generalization. Also these approaches are designed for simple environments such as gridworld and 2D bipedal terrains that are fully specified using a set of parameters, whereas our approach addresses more complex environments in the form of images and 3D objects that cannot be parameterized simply.
% \allen{PCGRL? procedural generation in games}

% \vspace{-3pt}
\textbf{Adversarial training with generative modeling.} Adversarial training \cite{goodfellow2014explaining} is popular in supervised learning (especially image classification) for improving the robustness of classifiers. One related direction shows that synthesizing adversarial data by searching over the latent space of a generative model can be more effective in attacking the classifier than searching over the raw image space \cite{jalal2017robust}. More recent work learns possible perturbations from pairs of datasets \cite{robey2020model} or pairs of original and perturbed data \cite{wong2020learning}. A closely related work is \cite{zakharov2019deceptionnet}, where a set of image perturbations are pre-specified and the model learns to be robust to confusing images through a minimax objective. Among other applications, \cite{lee2020shapeadv} attacks 3D point cloud classifiers by perturbing the latent variables of an autoencoder, similar to our setup. However, one key distinction is that while the loss/cost is differentiable through the classifier in supervised learning, the cost of an environment in our approach is determined by non-differentiable simulation. We resort to learning a differentiable cost predictor as a proxy. Adversarial training has also been applied to robotic grasping, either by randomly perturbing mesh vertices or training a Generative Adversarial Network (GAN) \cite{wang2019adversarial, morrison2020egad}. However, the difficulty of the objects is determined by a heuristic approach based on the antipodal metric of sampled grasps, unlike training a policy and evaluating the cost as in our approach.

% \vspace{-7pt}
\section{Problem Formulation}
\label{sec:formulation}
% \vspace{-7pt}
We assume that the discrete-time dynamics of the robot are given by $s_{t+1} = f_E(s_t, a_t)$ where $s_t \in \mathcal{S} \subseteq \mathbb{R}^{n_s}$  is the state of the robot at time-step $t$, $a_t \in \mathcal{A} \subseteq \mathbb{R}^{n_a}$ is the action, and $E \in \mathcal{E}$ is the environment that the robot is operating in. ``Environment" here broadly refers to external factors such as the object that a manipulator is trying to grasp, or the visual backdrop for a vision-based control task. Importantly, we do not assume knowledge of $\mathcal{E}$, which may be extremely high-dimensional and may not be parameterized simply. We assume access to a dataset $S := \{E_1, \dots, E_M\}$ of $M$ training environments. We let $P_0 := \sum_{i=1}^M \frac{1}{M} \delta_{E_i}$ denote the empirical distribution supported on $S$, where each $\delta_{E_i}$ is a Dirac delta distribution on $E_i$.

We assume that the robot has a sensor which provides observations $o_t \in \mathcal{O}$ of the environment. Let $\pi_\theta: \mathcal{O} \rightarrow \mathcal{A}$ denote a policy parameterized by $\theta \in \Theta$ (e.g., weights of a neural network). The robot's task is specified via a cost function; we let $C_E(\pi_\theta)$ denote the cumulative cost (over a specified time horizon) incurred by the policy $\pi_\theta$ when deployed in an environment $E$.
Our goal is to learn a policy $\pi_\theta$ that minimizes the \emph{worst-case expected cost} across a set $\mathcal{P}$ of distributions that contains $P_0$:
\begin{equation}
    \theta^\star = \underset{\Theta}{\newarginf} \ \underset{P \in \mathcal{P}}{\newsup} \ \underset{E \sim P}{\mathbb{E}} [C_E(\pi_\theta)].
\label{eq:minimax-goal}
\end{equation}

In the subsequent sections, we will demonstrate how to tackle the two main challenges highlighted in Section \ref{sec:intro}: (1) choosing a meaningful set $\mathcal{P}$ of distributions over which the worst-case optimization is performed, and (2) performing the inner maximization (generating meaningful, adversarial distributions over environments) for the outer minimization (training the policy). 

% \vspace{-7pt}
\section{Approach}
\label{sec:approach}
% \vspace{-7pt}

The overall approach is visualized in Fig.~\ref{fig:anchor}.
The key idea is to learn a generative model whose latent variables are cost-predictive and capture realistic variations in environments. This allows us to define a set $\mathcal{P}$ of distributions on this space for capturing realistic and task-relevant distribution shifts. We perform distributionally robust optimization using this set $\mathcal{P}$.

% \vspace{-10pt}
\subsection{Learning realistic variations in environments}
% \vspace{-5pt}
\label{sec:realistic}

Defining the uncertainty set $\mathcal{P}$ requires first defining the space over which distributions are supported. One option is to use the space of raw observations of environments (e.g., the raw pixel space of images).
However, this space can be extremely high-dimensional and perturbations in this space typically do not correspond to realistic variations in environments, which is evident in ``imperceptible attacks" in the image classification domain \cite{goodfellow2014explaining}. Instead, we opt for the latent space $\mathcal{Z} \subseteq \mathbb{R}^{n_z}$ of a generative model that captures realistic variations of environments. Previous work has demonstrated that perturbation or interpolation in the latent space can generate realistic variations in data \cite{joshi2019semantic, park2019deepsdf, achlioptas2018learning}. In the two examples detailed in Sec.~\ref{sec:experiments}, the raw environment is either a high-dimensional RGB image or a 3D object mesh. We use an autoencoder \cite{hinton2006reducing} as the generative model (Fig.~\ref{fig:model}):
\begin{equation}
    z = g(E), E' = f(g(E)),
\end{equation}
where the encoder $g: \mathcal{E} \rightarrow \mathcal{Z}$ maps the environment to a latent representation, and the decoder $f: \mathcal{Z} \rightarrow \mathcal{E}$ reconstructs the environment using the latent variable. Strictly speaking, an autoencoder is not a generative model but rather a representation model; however in practice, perturbations in the latent space generate meaningful variations in environments as shown in Sec.~\ref{sec:experiments} and in \cite{park2019deepsdf, achlioptas2018learning}. 

We embed the empirical distribution $P_0$ corresponding to the training dataset $S$ of environments (ref. Sec.~\ref{sec:formulation}) into the latent space $\mathcal{Z}$. This induces a distribution  $P_{\mathcal{Z}_0}$ on the latent space: 
\begin{equation}
% \color{blue}
P_{\mathcal{Z}_0} := \textstyle \sum_{i=1}^M \frac{1}{M} \delta_{z_{0_i}}, z_{0_i} = g(E_i), E_i \in S.
\end{equation}
We then define the set $\mathcal{P}$ using the \emph{Wasserstein distance} from optimal transport \cite{villani2009optimal}. For probability measures $X$ and $Y$ supported on $\mathcal{Z}$, and their couplings $\Pi(X,Y)$, the Wasserstein distance over the metric space $\mathcal{Z}$ is defined as:
\begin{equation}
% \color{blue}
W_d(X,Y) := \inf_{H \in \Pi(X,Y)} \mathbb{E}_H[d(x,y)],
\end{equation}
where $d(\cdot, \cdot)$ is the metric on the space $\mathcal{Z}$ (we use $d(x,y) = \|x-y\|_2$).
% denotes the nonnegative, symmetric transportation cost between two support points (we use $d(x,y) = \|x-y\|_2$).
We define the uncertainty set $\mathcal{P}$  as a Wasserstein ``ball'' around $P_{\mathcal{Z}_0}$ with radius $\rho$:
\begin{equation}
% \color{blue}
    \mathcal{P} := \{P_\mathcal{Z}: W_d(P_\mathcal{Z}, P_{\mathcal{Z}_0}) \leq \rho\}.
\end{equation}
Intuitively, the Wasserstein distance (also known as the ``earth mover's distance") measures the minimum cost of morphing one distribution into the other. There are two key advantages of the Wasserstein distance over other divergences (e.g., the KL-divergence or other $f$-divergences) that make it appealing in our setting. First, the Wasserstein distance is easier to interpret and more physically meaningful since it takes into account the underlying geometry of the space on which the probability distributions are defined (see, e.g., \cite{arjovsky2017wasserstein}). For example, consider distributions over objects only differing in their lengths and embedded in a latent space perfectly based on length (m). Consider three uniform distributions $P_{\mathcal{Z}_1}, P_{\mathcal{Z}_2}, P_{\mathcal{Z}_3}$ with supports $[0,1]$, $[1,2]$, and $[2,3]$ respectively. The Wasserstein distance captures our intuition that the distance between $P_{\mathcal{Z}_1}$ and $P_{\mathcal{Z}_2}$ is the same as that between $P_{\mathcal{Z}_2}$ and $P_{\mathcal{Z}_3}$ while the distance between $P_{\mathcal{Z}_1}$ and $P_{\mathcal{Z}_2}$ is smaller than that between $P_{\mathcal{Z}_1}$ and $P_{\mathcal{Z}_3}$.
The Wasserstein distance thus ensures distances in the latent embedding space correspond to differences in physical properties of the environments. Second, the Wasserstein distance between two distributions that do not share the same support can still be finite. This allows us to define a ball $\mathcal{P}$ around the distribution $P_{\mathcal{Z}_0}$ which contains distributions with differing supports; performing worst-case optimization over $P_{\mathcal{Z}_0}$ can thus provide robustness to unseen data.

% \vspace{-10pt}
\subsection{Learning task-relevant variations in environments}
\label{sec:task-relevant}
% \vspace{-5pt}

Ideally, we would like the set $\mathcal{P}$ to contain distributions over both realistic and \emph{task-relevant} features of environments. For example, consider a robotic manipulator learning to grasp objects and suppose that $\mathcal{P}$ contains different distributions over colors of objects. Such a set $\mathcal{P}$ can be very ``large" (in terms of the radius of the Wasserstein ball); however, performing distributionally robust optimization over $\mathcal{P}$ will result in a policy that is only robust to the task-irrelevant feature of color and not necessarily robust to task-relevant geometric features. Our key intuition is that the latent space $\mathcal{Z}$ captures task-relevant variations in the environments if (i) $\mathcal{Z}$ is \emph{cost-predictive} and (ii) closeness in the latent space corresponds to closeness in terms of costs. Then $\mathcal{P}$ captures cost-relevant (i.e., task-relevant) variations in environments. 

% {\color{blue}
By ``cost-predictive'', we mean that for the latent space $\mathcal{Z}$, there exists a mapping from the latent variable of the environment to the true cost of the environment. To satisfy (ii), such mapping should be Lipschitz continuous.
% latent space $\mathcal{Z}$ captures useful features that can predict the true cost of the environment it embeds.
%  since the space should relate the distance between two latent presentations to the difference in true costs of the actual environments,
\begin{definition}[$\gamma$-cost-predictive]
The latent space $\mathcal{Z}$ is $\gamma$-cost-predictive if there exists a $\gamma$-Lipschitz-continuous function $h$ that maps the latent variable of an environment to the true cost of the environment.
\end{definition}
% }

\begin{wrapfigure}[19]{l}{0.25\textwidth}
\vspace{-17pt}
\begin{center}
\includegraphics[width=0.25\textwidth]{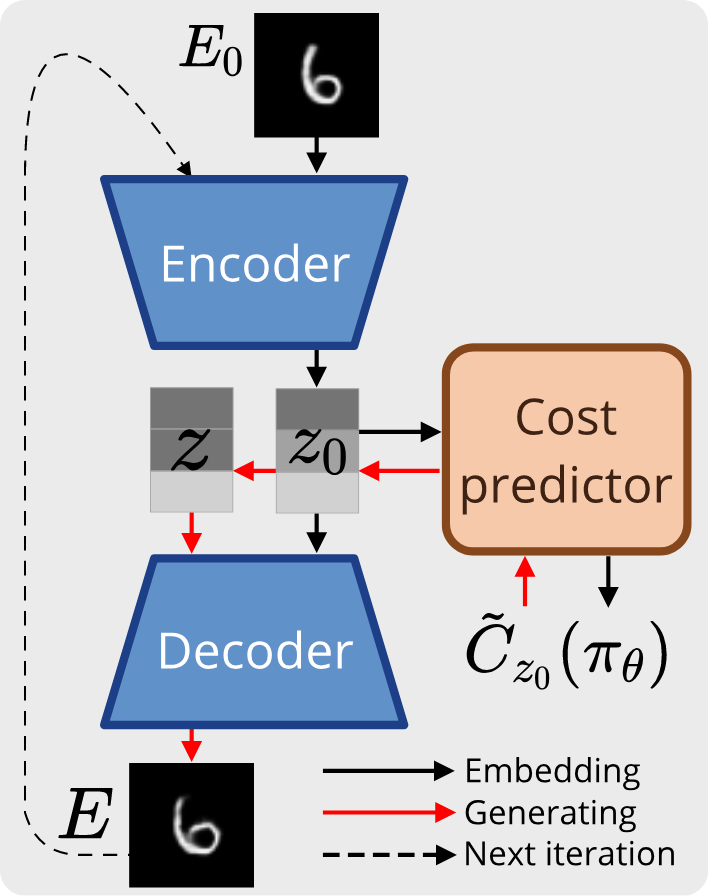}
\end{center}
\vspace{-8pt}
\caption{Training an autoencoder and a cost predictor allows iteratively generating adversarial environments (digit images used as visual backdrops) via gradient ascent on the latent space.}
\label{fig:model}
\end{wrapfigure}

In practice, we train a cost predictor $h_\psi: \mathcal{Z} \rightarrow \mathbb{R}$ that maps the latent variable $z$ of an environment $E$ to a predicted cost $\tilde{C}_z(\pi_\theta)$ 
% {\color{blue}
(the tilde denotes \emph{predicted} cost instead of true cost)
% }
, where $\pi_\theta$ is our current policy (Fig.~\ref{fig:model}). 
% {\color{blue}
The training labels are the true cost of the environments evaluated in simulation with the current policy, $C_E(\pi_\theta)$
% }
.
Furthermore, we constrain the Lipschitz constant $\gamma_{h_\psi}$ of the cost predictor. Now, environments with similar predicted costs are close in the latent space, and a Wasserstein ball with large radius contains distributions over environments with large variations in task-relevant features.
As described in Sec.~\ref{sec:optimization}, we iteratively update the policy by performing distributionally robust optimization via adversarial environment generation. Initially the policy may be sensitive to irrelevant features (e.g., color) and the set $\mathcal{P}$ contains distributions over irrelevant features. However, once we perform distributionally robust optimization with respect to $\mathcal{P}$, the policy becomes less sensitive to these irrelevant features, and $\mathcal{P}$ starts to capture task-relevant features (Fig.~\ref{fig:pendulum-envs}c). Without regularizing Lipschitzness, the latent space may be cost-predictive but far away points may have similar costs; in this case, the Wasserstein distance may not capture how much distributions differ in terms of task-relevant features (see ablation in Sec.~\ref{sec:experiments}).

\textbf{Embedding training.} In our experiments, we use a cost predictor $h_\psi$ with two linear layers and sigmoid activation; then $\gamma_{h_\psi}$ can be upper bounded \cite{sinha2017certifying}:
\begin{equation}
    \gamma_{h_\psi} \leqslant \overline{\gamma}_{h_\psi} := \|\psi_0\|_2 \|\psi_1\|_2 /16,
\label{eq:lip-bound}
\end{equation}
where $\psi_0$ and $\psi_1$ are the weight matrices at the two layers. In practice, we constrain $\overline{\gamma}_{h_\psi}$ to some fixed value $\overline{\gamma}$. Overall, we train the encoder $g$, decoder $f$, and cost predictor $h_\psi$ concurrently. The total loss function $\mathcal{L}$ is a weighted sum of four components:
\begin{equation}
\label{eq:embedding-loss}
    \mathcal{L} = \mathcal{L}_\text{rec} + \alpha_1 \mathcal{L}_{\text{pred}} + \alpha_2 \mathcal{L}_{\text{Lip}} + \alpha_3 \mathcal{L}_{\text{norm}}.
\end{equation}
where $\mathcal{L}_{\text{rec}}$ is the reconstruction loss of the autoencoder, $\mathcal{L}_{\text{pred}}$ is the $l_2$ loss between the predicted cost and true cost evaluated with the current policy, $\mathcal{L}_{\text{Lip}}$ is the $l_2$ loss between $\overline{\gamma}_{h_\psi}$ and $\overline{\gamma}$, and $\mathcal{L}_{\text{norm}}$ minimizes the norm of the embedded latent variables, which prevents the Lipschitz constant from being trivially constrained (by scaling the magnitude of latent variables to some fixed range).

% \vspace{-10pt}
\subsection{Distributionally Robust Policy Learning via Adversarial Environment Generation}
\label{sec:optimization}
% \vspace{-5pt}
% {\color{blue}
Next we explain our procedures for solving the minimax optimization \eqref{eq:minimax-goal}. From Sec.~\ref{sec:realistic}, we have chosen the uncertainty set as $\mathcal{P} = \{P_\mathcal{Z}: W_d(P_\mathcal{Z}, P_{\mathcal{Z}_0}) \leq \rho\}$. The optimization problem (subject to the uncertainty set constraint) can be re-formulated as:
% which is reiterated below:
\begin{equation}
    \minimize_{\pi_\theta} \ \underset{P_ \mathcal{Z}}{\newsup} \ \underset{z \sim P_ \mathcal{Z}}{\mathbb{E}} [C_{f(z)}(\pi_\theta)],
\label{eq:drpl-minimax}
\end{equation}
% since this entails performing worst-case optimization over distributions
where $f(z)$ is the reconstructed environment (passing $z$ through the decoder $f$). Searching over $\mathcal{P}$ exactly is intractable; we thus follow \cite{sinha2017certifying} by applying a Lagrangian relaxation with a penalty parameter $\lambda \geq 0$:
\begin{equation}
    \minimize_{\pi_\theta} \ \sup_{P_\mathcal{Z}} \ \{\mathbb{E}_{P_\mathcal{Z}} [C_{f(z)}(\pi_\theta)]-\lambda W_d(P_{\mathcal{Z}},P_{\mathcal{Z}_0}) \},
\label{eq:drpl-lagrangian}
\end{equation}
% Note that now the supremum is over the latent distribution $P_\mathcal{Z}$ now instead of the original distribution of environments $P$. 

Maximizing over the distribution $P_\mathcal{Z}$ is still difficult, and thus we further apply an equivalent dual re-formulation \cite{sinha2017certifying} to allow maximizing over the latent variable $z$ instead:
\begin{equation}
    \minimize_{\pi_\theta} \ \underset{z_0 \sim P_{\mathcal{Z}_0}}{\mathbb{E}} \{\sup_z \ [C_{f(z)}(\pi_\theta) - \lambda d(z,z_0)] \}.
\label{eq:drpl-dual}
\end{equation}
% where $f(z)$ is the reconstructed new environment and . 

Equivalence between \eqref{eq:drpl-lagrangian} and \eqref{eq:drpl-dual} requires $C_{f(z)}(\pi_\theta)$ to be continuous in $f(z)$ \cite{sinha2017certifying}, which is not reasonable as the space $\mathcal{E}$ where $f(z)$ belongs can be extremely high-dimensional (e.g.~pixel space for images), and $C_{f(z)}(\pi_\theta)$ is evaluated using non-differentiable simulation.
To eschew the issue, we substitute $C_{f(z)}(\pi_\theta)$ with the predicted cost $\tilde{C}_z(\pi_\theta)$ from the $\overline{\gamma}$-Lipschitz cost predictor,
\begin{equation}
    \minimize_{\pi_\theta} \ \underset{z_0 \sim P_{\mathcal{Z}_0}}{\mathbb{E}} \{\sup_z \ [\tilde{C}_z(\pi_\theta) - \lambda d(z,z_0)] \},
\label{eq:drpl-final-app}
\end{equation}
% This re-formulation turns the problem of maximizing over distributions to one of maximizing over the latent variable $z$.
% (see App. A1 of the extended version of the paper \cite{ren2021distributionally})
% We approximate the true cost using the predicted cost, and 
% ; see App. A2 of \cite{ren2021distributionally} for the detailed algorithm)
Now the inner supremum can be performed with gradient ascent on the latent space (Fig.~\ref{fig:model}:
\begin{equation}
\label{eq:gradient-ascent}
    z \leftarrow z_0 + \eta \nabla_z[\tilde{C}_{z}(\pi_\theta) - \lambda d(z,z_0)],
\end{equation}
where $\eta$ is the step size. In practice we perform the minimax procedure iteratively: during inner maximization, a set of $K$ latent variables $\{z_{0_i}\}_{i=1}^K$ are sampled from the current latent distribution $P_{\mathcal{Z}_0}$ and perturbed into $\{z_i\}_{i=1}^K$. The reconstructed environments $\{E_i\}_{i=1}^K := \{f(z_i)\}_{i=1}^K$ are added to the dataset $S$; during the minimization phase, the policy is re-trained using the augmented $S$. Between the two phases, we train the embedding using all environments in $S$ and their cost evaluated at the current policy $\pi_\theta$; this is used to update $P_{\mathcal{Z}_0}$, whose support grows over iterations.
% }
\vspace{-5pt}
% \begin{wrapfigure}{L}{0.5\textwidth}
% \begin{minipage}{0.5\textwidth}
\begin{algorithm}[H]
\caption{Distributionally Robust Policy Learning via Adversarial Environment Generation}
\begin{algorithmic}[1]
    \Require $S$, initial set of environments; $\pi_\theta$, initial policy
    \State Pre-train $\pi_\theta$ with $S$
    \For{$N$ iterations} \Comment{Run minimax $N$ times}
        \State Evaluate $C_E(\pi_\theta), \forall E \in S$ 
    	\State Train embedding with \eqref{eq:embedding-loss} and update $P_{\mathcal{Z}_0}$
        \State Sample $\{z_{0_i}\}_{i=1}^K \sim P_{\mathcal{Z}_0}$; generate $\{E_i\}_{i=1}^K$ with \eqref{eq:gradient-ascent} and add to $S$
        \State Improve $\pi_\theta$ with $S$
    \EndFor
	\end{algorithmic}
\normalsize
\end{algorithm}
% \end{minipage}
% \end{wrapfigure}
\vspace{-10pt}
% {\color{blue}
\textbf{Target ascent in the latent space.} In practice we find it difficult to tune the number of gradient ascent steps when perturbing a latent variable. Instead, we set a target on how much the predicted cost $\tilde{C}_{z}(\pi_\theta)$ of the perturbed latent variable should increase from that of the original latent variable $\tilde{C}_{z_0}(\pi_\theta)$, and run \eqref{eq:gradient-ascent} until the target is reached. However, since the cost predictor is Lipschitz-regularized, the range of its output is likely to be smaller than the true range of the cost evaluated in simulation. Thus at each iteration before generating new environments, we calculate the empirical range $\mathcal{R}(\tilde{C})$ (difference of the maximum and minimum predicted cost over all environments in $S$), and set the target ascent $\Delta \tilde{C}$ using a percentage $\Delta \tilde{C}_p \in [0,1]$ of $\mathcal{R}(\tilde{C})$. We find that $\Delta \tilde{C}_p = 0.2$ works well for all experiments.
%  (i.e.,~the range of labels used for training the cost predictor)
% \textbf{Distributionally robust bound on the predicted cost.} We end by deriving a robustness result that follows from Kantorovich-Rubinstein duality \cite{villani2009optimal} of the Wasserstein distance (see App. A3 of \cite{ren2021distributionally}). The following result bounds the worst-case expected predicted cost of a learned policy $\pi_\theta$ over the set $\mathcal{P}$ of distributions in terms of the (predicted) cost on the empirical distribution $P_{\mathcal{Z}_0}$.
% % \vspace{-5pt}
% \begin{theorem}
% \label{thm:generalization-bound} For the cost predictor $h_\psi$ with Lipschitz constant upper bounded by $\overline{\gamma}$, and the uncertainty set $\mathcal{P} := \{P_\mathcal{Z}: W_d(P_\mathcal{Z}, P_{\mathcal{Z}_0}) \leq \rho\}$, the following inequality holds:
% \begin{equation}
%     \sup_{P_\mathcal{Z} \in \mathcal{P}} \underset{z \sim P_\mathcal{Z}}{\mathbb{E}}[\tilde{C}_{z}(\pi_\theta)] & \leq  \underset{z \sim P_{\mathcal{Z}_0}}{\mathbb{E}}[\tilde{C}_{z}(\pi_\theta)] + \overline{\gamma} \rho.
% \label{eq:generalization-bound}
% \end{equation}
% % \vspace{-10pt}
% \end{theorem}
% }
% \vspace{-8pt}
\section{Experiments}
\label{sec:experiments}
% \vspace{-7pt}

% \vspace{-10pt}
% \begin{figure*}[t]
% \begin{center}
%   \begin{minipage}[c]{0.85\textwidth}
%     \includegraphics[width=\textwidth]{figures/pendulum_envs_comparison.jpg}
%   \end{minipage}\hfill
% % \vspace{-3pt}
% \caption{(a) Landmark, (b) Digit, and (c) Urban images used in pendulum task. Top right shows the robot view from the onboard camera. (d) Original and perturbed landmark images (cropped) at different iterations. Initially generated images contain both task-relevant (landmark dimension/location) and irrelevant (landmark/background color) variations, but after iterations only task-relevant variations remain.}
% \label{fig:pendulum-envs}
% \end{center}
% \vspace{-23pt}
% \end{figure*}

We implement our approach on two robotics tasks in simulation: (1) swinging up a pendulum with onboard vision, and (2) grasping realistic 3D objects. We also test grasping policies on a real robot arm. Through these experiments we aim to investigate the following questions: (1) Does our method offer superior OoD performance compared to data augmentation or domain randomization techniques? (2) Does our method generate seemingly meaningful environments for training? (3) Does regularizing the Lipschitz constant of the cost predictor lead to more meaningful environment variations and better OoD performance? (4) Does our method improve sim2real performance for the grasping task? For all experiments in simulation we run the minimax procedures for 30 iterations, and all results are evaluated at the iteration with the best training performance and averaged over 10 seeds. See App. A4/A5 of the extended version \cite{ren2021distributionally} for more ablation studies, experimental details, and hyperparameters.

\subsection{Swinging up a pendulum with onboard vision}
\begin{figure}
\includegraphics[width=0.49\textwidth]{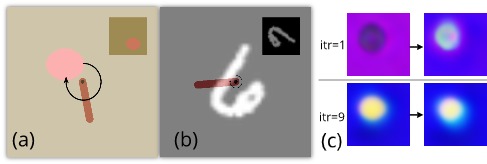}
\caption{(a) Landmark and (b) Digit images used in pendulum task. Top right shows the robot view from the onboard camera. (c) Original and perturbed landmark images (cropped) at different iterations. Initially generated images contain both task-relevant (landmark dimension/location) and irrelevant (landmark/background color) variations, but after iterations only task-relevant variations remain.}
\label{fig:pendulum-envs}
\vspace{-15pt}
\end{figure}

% \vspace{-5pt}
\textbf{Task and environment specification.} Imagine a camera mounted to a pendulum and facing a visual backdrop (Fig.~\ref{fig:pendulum-envs}); the pendulum needs to swing up and balance itself using visual feedback. This is different from typical image-based pendulum tasks where the virtual camera is located away from the pendulum and is pointed at the rotating pendulum and a static backdrop (distraction). Our onboard camera setup is more representative of robotics tasks (e.g., vision-based navigation) and requires the policy to extract features from the backdrop. We consider an environment $E$ as a backdrop image, and use two types of images (Fig.~\ref{fig:pendulum-envs}): (1) Landmark: randomly colored backdrop with a randomly colored, elliptical ``landmark'' at a fixed radial location; (2) Digit: black backdrop with white digits from the MNIST \cite{lecun1998gradient} and USPS \cite{hull1994database} datasets. At each time-step, the robot's policy maps image observations from the past three time-steps to the torque applied at the joint.
% ; (2) Urban: photo-realistic urban environment images from the Apollo Synthetic dataset \cite{ap}. 

\textbf{Control policy training.} We perform off-policy training using Soft Actor Critic (SAC) \cite{haarnoja2018soft}. Episodes are sampled with the pendulum initialized at any angle. The reward function penalizes angle deviation from upright, angular velocity, and torque applied.
% (same as pendulum task from OpenAI Gym suite \cite{brockman2016gym}).
% ; empirically, on-policy policy gradient algorithms do not converge well for the task. 
% \vspace{-1pt}

\textbf{DRAGEN training.} The training dataset of images is embedded in the low-dimensional latent vector space of an autoencoder. Both the encoder and decoder consist of convolutional layers and linear layers, and the decoder upsamples bilinearly. For training the cost predictor, we evaluate the cost of each image using average cost of episodes with the pendulum initialized around the lowest point. The cost is normalized between $[0,1]$; the lower bound corresponds to the pendulum not moving at all with itself hanging downwards, and the upper bound corresponds to the cost when the policy is trained using the true states of the pendulum instead of the camera image. 
% using estimates of the lower and upper bound.

% \vspace{-1pt}
\textbf{Baselines.} We benchmark DRAGEN against commonly-used data augmentation techniques in RL including pixel-wise Gaussian noise, Perlin noise \cite{perlin1985image}, and random cutout \cite{laskin2020reinforcement}. Note that since the policy needs to extract spatial information from the image, some other techniques such as flipping and rotating cannot be applied.

\setlength{\tabcolsep}{4pt} % Make spacing between table columns smaller; default value: 6pt
\begin{table*}[h!]
% \scriptsize
\footnotesize
% \vspace{-5pt}
\begin{center}
\centerline{    % adds this makes it actually centered
\begin{tabular}{cccccccc}
    \toprule
    & \phantom{a} & Train & \phantom{a} & \multicolumn{4}{c}{Test}\\
    \cmidrule{3-3} \cmidrule{5-8}
    {Method} & &{Normal} && {Closer} & {Farther} & {Smaller} & {Larger} \\ \midrule
    DRAGEN && 0.858 $\pm$ 0.026 && \bftab 0.432 $\pm$ 0.018 & \bftab 0.577 $\pm$ 0.024 & \bftab 0.761 $\pm$ 0.025 & \bftab 0.740 $\pm$ 0.016 \\
    Perlin noise && 0.798 $\pm$ 0.031 && 0.350 $\pm$ 0.027 & 0.495 $\pm$ 0.033 & 0.638 $\pm$ 0.032 & 0.678 $\pm$ 0.024 \\
    Gaussian noise && 0.892 $\pm$ 0.030 && 0.357 $\pm$ 0.023 & 0.499 $\pm$ 0.025 & 0.699 $\pm$ 0.028 & 0.681 $\pm$ 0.026 \\
    None && \bftab 0.892 $\pm$ 0.022 && 0.359 $\pm$ 0.023 & 0.517 $\pm$ 0.028 & 0.714 $\pm$ 0.025 & 0.669 $\pm$ 0.026 
    \\
    \bottomrule
\end{tabular}}
\vspace{2pt} 
% \newline
\centerline{
\begin{tabular}{cccccccc}
    & \multicolumn{6}{c}{Test}\\
    \cmidrule{2-8}
    {Method} & 2 & 3 & 4 & 5 & 6 & 7 & 9\\ \midrule
    DRAGEN & \bftab 0.608 $\pm$ 0.016 & 0.787 $\pm$ 0.021 & \bftab 0.626 $\pm$ 0.023 & \bftab 0.589 $\pm$ 0.018 & \bftab 0.842 $\pm$ 0.012 & \bftab 0.841 $\pm$ 0.014 & \bftab 0.793 $\pm$ 0.018\\
    DRAGEN-NoLip & 0.519 $\pm$ 0.022 & 0.771 $\pm$ 0.023 & 0.579 $\pm$ 0.025 & 0.541 $\pm$ 0.020 & 0.789 $\pm$ 0.018 & 0.782 $\pm$ 0.014 & 0.654 $\pm$ 0.040 \\
    DRAGEN-NoCost & 0.553 $\pm$ 0.020 & 0.785 $\pm$ 0.023 & 0.611 $\pm$ 0.025 & 0.522 $\pm$ 0.019 & 0.755 $\pm$ 0.020 & 0.802 $\pm$ 0.018 & 0.741 $\pm$ 0.022 \\
    Perlin noise & 0.540 $\pm$ 0.021 & \bftab 0.797 $\pm$ 0.018 & 0.618 $\pm$ 0.028 & 0.529 $\pm$ 0.021 & 0.750 $\pm$ 0.023 & 0.814 $\pm$ 0.019 & 0.732 $\pm$ 0.025 \\
    Gaussian noise & 0.551 $\pm$ 0.018 & 0.754 $\pm$ 0.026 & 0.613 $\pm$ 0.027 & 0.511 $\pm$ 0.021 & 0.679 $\pm$ 0.027 & 0.772 $\pm$ 0.022 & 0.707 $\pm$ 0.031 \\
    Random cutout & 0.549 $\pm$ 0.023 & 0.779 $\pm$ 0.016 & 0.609 $\pm$ 0.025 & 0.543 $\pm$ 0.019 & 0.756 $\pm$ 0.020 & 0.772 $\pm$ 0.021 & 0.648 $\pm$ 0.040 \\
    None & 0.551 $\pm$ 0.027 & 0.761 $\pm$ 0.020 & 0.552 $\pm$ 0.037 & 0.540 $\pm$ 0.023 & 0.733 $\pm$ 0.016 & 0.803 $\pm$ 0.017 & 0.724 $\pm$ 0.034 \\
    \bottomrule
\end{tabular}}
% \vspace{2pt} \newline
% \begin{tabular}{SSSSSSSS}
%     & \phantom{a} & Train & \phantom{a} & \multicolumn{4}{c}{Test}\\
%     \cmidrule{3-3} \cmidrule{5-8}
%     {Method} && {9am Clear} && {9am Light Rain} & {9am Heavy Rain} & {5pm Clear} & {6pm Clear} \\ \midrule
%     DRAGEN && 0.776 $\pm$ 0.027 && \bftab 0.553 $\pm$ 0.025 & 0.546 $\pm$ 0.030 & \bftab 0.470 $\pm$ 0.027 & \bftab 0.439 $\pm$ 0.031 \\
%     Perlin noise && 0.793 $\pm$ 0.026 && 0.540 $\pm$ 0.028 & 0.509 $\pm$ 0.032 & 0.418 $\pm$ 0.031 & 0.286 $\pm$ 0.040 \\
%     Gaussian noise && \bftab 0.801 $\pm$ 0.022 && 0.546 $\pm$ 0.027 & \bftab 0.552 $\pm$ 0.029 & 0.411 $\pm$ 0.034 & 0.215 $\pm$ 0.046 \\
%     Random cutout && 0.772 $\pm$ 0.035 && 0.481 $\pm$ 0.030 & 0.470 $\pm$ 0.031 & 0.402 $\pm$ 0.029 & 0.275 $\pm$ 0.035 \\
%     None && 0.793 $\pm$ 0.023 && 0.516 $\pm$ 0.032 & 0.508 $\pm$ 0.038 & 0.290 $\pm$ 0.040 & 0.257 $\pm$ 0.041 \\
%     \bottomrule
% \end{tabular}
% \vspace{5pt}
\caption{Normalized reward (mean and standard deviation over 10 seeds) for the pendulum task. Top: Landmark; bottom: Digit.}
\label{tab:pendulum}
\end{center}
\vspace{-20pt}
\end{table*}

\textbf{Results: Landmark.} We generate a set of training environments (``Normal'' in Table \ref{tab:pendulum}, 200 images) where the landmarks are centered along the radial direction and normally sized, and four sets of test environments where the landmarks are closer to or farther away from the center, or smaller or larger in dimensions (``Closer'', ``Farther'', ``Smaller'', ``Larger'' in Table \ref{tab:pendulum}). DRAGEN outperforms all baselines among all test datasets. Fig.~\ref{fig:pendulum-envs}c demonstrates that DRAGEN learns to focus on generating task-relevant variations such as landmark locations and dimensions over iterations.

\begin{figure}
\includegraphics[width=0.49\textwidth]{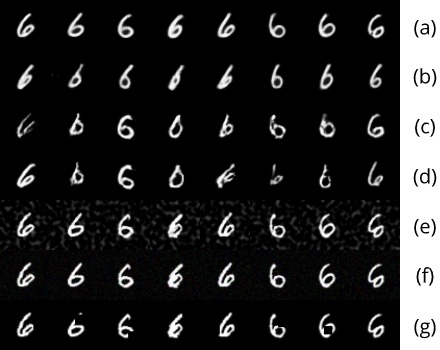}
\caption{Samples of images of digit 6 at one iteration: (a) original, (b) DRAGEN, (c) DRAGEN-NoLip, (d) DRAGEN-NoCost, (e) Perlin noise \cite{perlin1985image}, (f) pixel-wise Gaussian noise, and (g) random cutout. Images at the same column are based on the same original image on the top row. DRAGEN generates new images that tend to rotate or straighten the long stroke of digit 6; such features are cost/task-relevant. Variations generated by DRAGEN-NoLip and DRAGEN-NoCost tend to be irregular and disorganized. Also Perlin noise tends to ``augment'' the white digit with its structure, and thus provides a strong baseline for the task.}
\label{fig:digit-comparison}
\vspace{-20pt}
\end{figure}

% \vspace{-1pt}
\textbf{Results: Digit.} We run experiments using digits ${2,3,4,5,6,7,9}$; digits ${0,1,8}$ are not suitable due to symmetry about both axes. Policies are trained separately for each digit with 200 training images from the MNIST dataset and then tested on the USPS dataset. The USPS dataset contains human drawings with more cursive digits. Due to space limits, we only show test performance in Table~\ref{tab:pendulum}. DRAGEN outperforms all baselines for digits ${2,5,6,7,9}$, and performs comparably to the strongest baseline for ${3,4}$. 
% We find that many digits ${3,4}$ from the USPS dataset are significantly different from those in MNIST dataset. 
Note that Perlin noise is a strong baseline for this setting as its structure may ``augment'' the white digit (Fig.~\ref{fig:digit-comparison}e). In App. A4 of \cite{ren2021distributionally}, we show that DRAGEN outperforms Perlin noise in larger margins when colored distractions are added to the black background.

% \textbf{Results: Urban.} The policy is trained using 270 images simulating urban scenes in clear weather at 9 am (``9am Clear'' in Table \ref{tab:pendulum}), and tested on four other settings varying in weather and time of day (Fig.~\ref{fig:anchor}). Images here are significantly more complex than landmark and digit images, and we find that DRAGEN struggles to generate fine-grained variations in images, limited by the simple generative modeling and small amount of data. Nonetheless, DRAGEN shows stronger OoD generalization in two settings (``5pm Clear'' and ``6pm Clear'') where the natural lighting is dimmer, while also matching the strongest baseline in rainy conditions.

% \vspace{-3pt}
\textbf{Ablation: Regularizing Lipschitz constant of the cost predictor.} We also investigate whether it is useful to constrain the Lipschitz constant of the cost predictor, which we hypothesize induces task-relevant variations in generated environments. We run the additional baseline without Lipschitz regularization (``DRAGEN-NoLip'' in Table \ref{tab:pendulum}) using digit datasets, and it performs worse than DRAGEN across all test datasets. Fig.~\ref{fig:digit-comparison}c shows that DRAGEN-NoLip generates less task-relevant variations in images.

\textbf{Ablation: Learning a cost-predictive embedding of environments.} We remove the cost prediction loss $L_\text{pred}$ from \eqref{eq:embedding-loss} - the cost predictor is not learned, and thus it is not possible to perform gradient ascent in the latent space to find adversarial environments. Instead, we randomly perturb the latent variables of existing environments in the latent space to generate new ones. Perturbations are sampled from zero-mean Gaussian distributions with diagonal covariances to roughly match the amount of perturbations generated by DRAGEN. The results are shown in Table.~\ref{tab:pendulum} as DRAGEN-NoCost. Without searching for adversarial environments varying in task-relevant features, the baseline performs worse than DRAGEN and on par with other data augmentation techniques. Fig.~\ref{fig:digit-comparison}d shows that DRAGEN-NoCost generates images with worse qualities than DRAGEN.

% {\color{blue}
\textbf{Runtime comparison.} Each experiment is run using one Nvidia RTX 2080Ti GPU and 16 server CPU threads. It takes 3 hours to run 30 iterations of DRAGEN or DRAGEN-NoLip training, and 2.5 hours for DRAGEN-NoCost. All other baselines take about 2 hours.
% }
% \vspace{-12pt}
\subsection{Grasping realistic 3D objects}
% \vspace{-2pt}

\begin{wrapfigure}[12]{r}{2.8cm}
\vspace{-18pt}
\begin{center}
    \begin{tabular}{@{}c@{}}
    \includegraphics[width=0.15\textwidth]{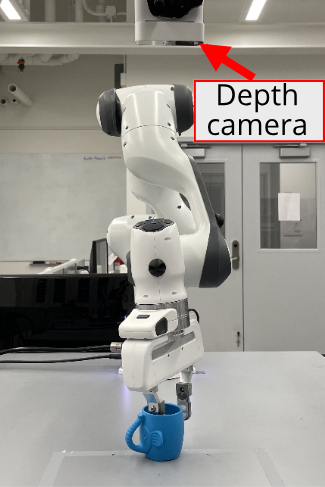} \\
    \end{tabular}
\end{center}
\vspace{-8pt}
\caption{Hardware setup for grasping.}
\label{fig:grasp-hardware}
\end{wrapfigure}

\textbf{Task and environment specification.} A robot arm needs to pick up an object placed at a fixed location in the PyBullet simulator \cite{coumans2021}. Before executing the grasp, the robot receives a heightmap image from an overhead camera and decides the 3D positions and yaw orientation of the grasp. We consider an environment $E$ as an object, and diverse synthetic 3D objects from the 3DNet \cite{wohlkinger20123dnet} dataset are used for training. Policies trained in simulation with synthetic data are also transferred to a real setup (Fig. \ref{fig:grasp-hardware}). 
% To simplify the task, objects are not colored and RGB image information is not used. 
% Two types of objects are used: (1) 2D objects: objects extruded from 2D primitives including rectangles, ellipses, and triangles; (2)
% and ShapeNetCore \cite{chang2015shapenet}
\begin{table*}[h!]
\footnotesize
% \scriptsize
\vspace{-5pt}
\begin{center}
% \centerline{    % adds this makes it actually centered
% \begin{tabular}{SSSSSSSSSS}
%     \toprule
%     & \phantom{a} & & \phantom{a} & \multicolumn{3}{c}{Test Success - 2D primitives} & \phantom{a} &\\
%     \cmidrule{5-7}
%     Method && Train Reward && 0.3 & 0.4 & 0.5 && Hardware \\ \midrule
%     DRAGEN  && 0.929 $\pm$ 0.008 && \bftab 0.655 $\pm$ 0.029 & \bftab 0.684 $\pm$ 0.024 & \bftab 0.716 $\pm$ 0.023 && \bftab 1.0, 0.9, 0.9 \\
%     DR     && 0.882 $\pm$ 0.034 && 0.606 $\pm$ 0.030 & 0.632 $\pm$ 0.031 & 0.672 $\pm$ 0.024 && 0.8, 0.8, 0.8 \\
%     None   && \bftab 0.948 $\pm$ 0.015 && 0.577 $\pm$ 0.036 & 0.627 $\pm$ 0.027 & 0.686 $\pm$ 0.025 && 0.8, 0.7, 0.7 \\
%     % \bottomrule
% \end{tabular}}
% \vspace{2pt} \newline
% \setlength{\tabcolsep}{2pt} % Default value: 6pt
\centerline{
\begin{tabular}{cccccccccc}
    \toprule
    % && {0.3} & {0.4} & {0.5}
    % & \multicolumn{3}{c}{Test Success - ShapeNetCore}
    & \phantom{a} & & \phantom{a} & \multicolumn{3}{c}{Test Success - 3DNet} & \phantom{a} & \\
    \cmidrule{5-7} 
    % \cmidrule{7-9}
    {Method} & \phantom{a} & {Train Reward} & \phantom{a} & {0.3} & {0.4} & {0.5} & \phantom{a} & Hardware \\ \midrule
    % & \bftab 0.508 $\pm$ 0.048 & \bftab 0.659 $\pm$ 0.039 & \bftab 0.695 $\pm$ 0.058
    DRAGEN && \bftab 0.911 $\pm$ 0.011 && \bftab 0.733 $\pm$ 0.025 & \bftab 0.877 $\pm$ 0.034 & \bftab 0.939 $\pm$ 0.024 && \bftab 0.975, 0.975, 0.95 \\
    % & 0.471 $\pm$ 0.046 & 0.613 $\pm$ 0.042 & 0.679 $\pm$ 0.052 &
    DR && 0.866 $\pm$ 0.030 && 0.723 $\pm$ 0.038 & 0.814 $\pm$ 0.035 & 0.861 $\pm$ 0.044 && 0.90, 0.90, 0.90 \\
    EGAD && 0.875 $\pm$ 0.016 && 0.721 $\pm$ 0.029 & 0.831 $\pm$ 0.040 & 0.853 $\pm$ 0.029 && 0.925, 0.90, 0.90 \\
    % & 0.505 $\pm$ 0.050 & 0.636 $\pm$ 0.048 & 0.688 $\pm$ 0.054 &
    None && 0.890 $\pm$ 0.034 && 0.703 $\pm$ 0.051 & 0.748 $\pm$ 0.045 & 0.813 $\pm$ 0.040 && 0.85, 0.85, 0.825 \\
    \bottomrule
\end{tabular}}
% \vspace{5pt}
\caption{Results (mean and standard deviation (over 10 seeds) for the grasping task.}
\label{tab:grasping}
\end{center}
\vspace{-25pt}
\end{table*}

\textbf{Control policy training.} We follow the off-policy Q learning from \cite{zeng2018learning}. The Q function is modeled as a fully convolutional network (FCN) that maps a heightmap image to a pixel-wise prediction of success of executing the grasp at the corresponding 2D location. The heightmap image is rotated into six different orientations (30 degree interval) and stacked as input to the network. The pixel with the highest value across the six output maps is used as the 2D position and yaw orientation of the grasp. The grasp height is chosen as $3$ centimeters lower than the height value at the picked pixel. The friction coefficient is fixed as $0.3$. The reward function is either 0 or 1 based on whether the object is successfully lifted. Due to the use of fully convolutional network and multiple grasp orientations, the cost of an object is invariant to its position and orientation.

\textbf{DRAGEN training.} The training dataset of object meshes is embedded in the low-dimensional latent vector space of an autoencoder. Generative modeling for 3D object meshes is more involved than that for images. The encoder is a PointNet network \cite{qi2017pointnet} that encodes objects from sampled 3D points on their surfaces. The decoder follows recent work in learning continuous signed distance functions (SDF, distance of a spatial point to the closest surface) for shape representation \cite{park2019deepsdf, kleineberg2020adversarial}: it maps the pair of a query 3D location and latent variable to the SDF value at that location. After querying the SDF at many points, the mesh can be rasterized via the Marching cubes algorithm.

Training the cost predictor requires a continuous cost for the objects to allow for gradient ascent in the latent space. We assign the cost of an object between $[0,1]$ based on the minimum friction coefficient among $[0.10,0.55]$ needed for a successful grasp (lower value corresponds to lower cost).
% 11 discrete levels from 0 to 1 with 0.1 interval, 0 for failed grasp at all coefficients, and other value corresponds to coefficient from 0.1 to 0.55 with 0.05 interval.

\begin{figure}
\includegraphics[width=0.45\textwidth]{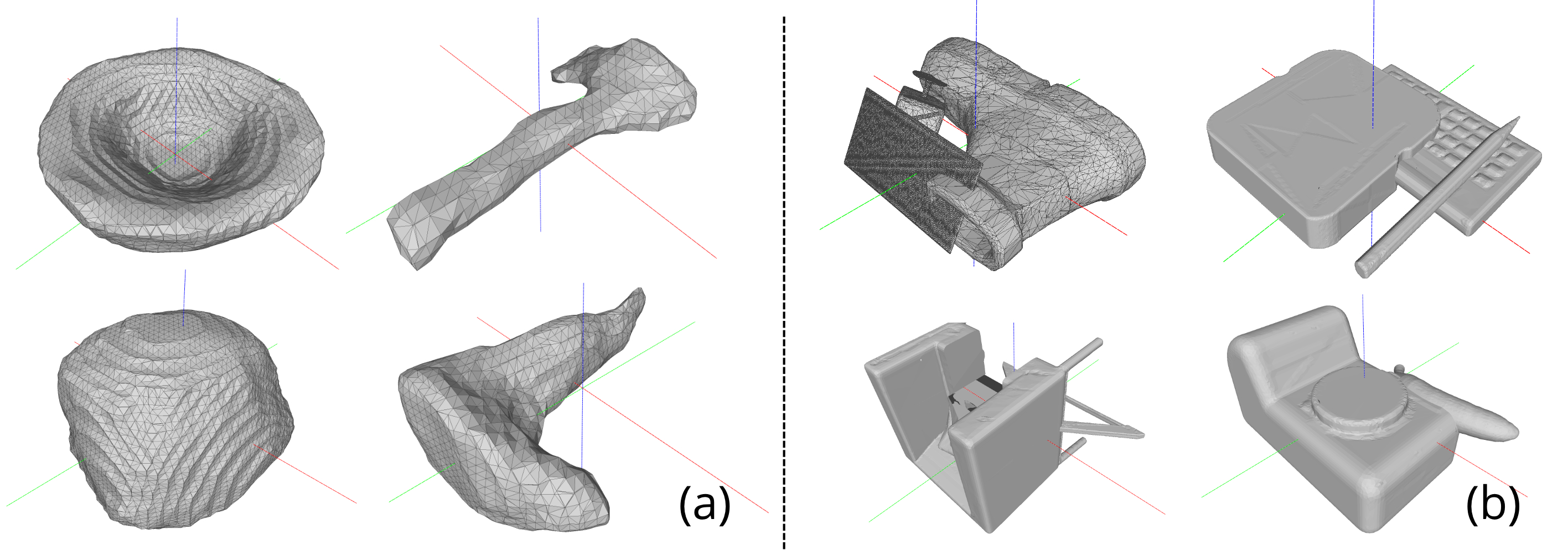}
\caption{Sample objects generated by (a) DRAGEN, (b) Domain Randomization. Objects in (b) are unrealistic and more irregular in shapes. Although they vary in shapes significantly, they are not realistic and can hinder the training progress. Objects generated by DRAGEN generally contain less perturbations from original objects and tend to be more realistic.}
\label{fig:meshes}
% \end{wrapfigure}
\vspace{-20pt}
\end{figure}

\textbf{Baselines.} Besides no data augmentation (None), we also implement Domain Randomization (DR) technique from \cite{tobin2018domain} that randomly chains shape primitives into new objects. We hypothesize that DR does not generate realistic objects and can be less efficient and effective when training policies to be tested on realistic objects. 
% {\color{blue}
Another baseline (EGAD) is to substitute adversarial objects generated at each iteration with objects from the EGAD dataset \cite{morrison2020egad}. The EGAD dataset consists of grasping objects of diverse complexities and difficulties, but most of them are not realistic, especially the ones with high complexity and difficulty.
% }

% {\color{red}We also compare DRAGEN to the EGAD dataset in Appendix \ref{app:ablation}}.

% \textbf{Results: 2D objects.} Training and test datasets are generated using two different distributions of parameters on the dimensions and shape of the objects (see details in App. A5 of \cite{ren2021distributionally}). While the friction coefficient is fixed in training, trained policies are tested at three different values $\{0.3, 0.4, 0.5\}$. Table \ref{tab:grasping} shows that DRAGEN outperforms both baselines in the test dataset, and the results are consistent across different friction coefficients.

\textbf{Results} The 60 categories of objects from the 3DNet dataset are split into training and test datasets, each with 255 and 205 objects. Table \ref{tab:grasping} shows that DRAGEN outperforms all baselines again in both test datasets. Surprisingly, DRAGEN also performs best in the training dataset. Our generated objects (Fig.~\ref{fig:meshes}) may form a better training curriculum than those generated with Domain Randomization or objects from the EGAD dataset. Both DR and EGAD baselines performed worse than no augmentation in training reward. 

\textbf{Results: Hardware.} Policies trained in simulation are tested with 40 common objects (see App. A5 of \cite{ren2021distributionally} for images and discussion) on a Franka Panda arm and a Microsoft Azure Kinect depth camera. For each method, we run 3 out of the 10 policies trained in simulation (corresponding to the 10 different seeds). DRAGEN performs the best in both settings (Table \ref{tab:grasping}). Videos of representative trials are provided in the supplementary materials.
% Policies for 2D objects are tested with 10 toy blocks in rectangular, elliptical, and triangular shapes, and 

% {\color{blue}
\textbf{Runtime comparison.} Each experiment is run using 1 RTX 2080Ti GPU and 32 server CPU threads. It takes about 6 hours to run 30 iterations of DRAGEN training.  The domain randomization baseline takes longer time (about 8 hours) since generating new objects involves chaining shape primitives and checking if all primitives overlap.  Without any data augmentation, the baseline takes about 4 hours to run. EGAD training takes the same time since new objects added to the dataset are pre-available.
% }

% \begin{figure}
% \centering
% \includegraphics[width=0.40\textwidth]{figures/3d_objects.jpg}
% \caption{40 common objects used in hardware experiments.}
% \label{fig:objects-hardware}
% \vspace{-20pt}
% \end{figure}

\vspace{-3pt}
\section{Conclusion}
\label{sec:conclusion}
\vspace{-1pt}
We have presented DRAGEN, a framework that iteratively improves the robustness of control policies to realistic distributional shifts. By training a generative model with a cost-predictive latent space, DRAGEN can generate task-relevant and realistic variations in environments, which are then added to the training dataset to improve the policy. Results on two different robotic tasks in simulation and in sim2real transfer demonstrate the strong OoD performance of our approach. 

\emph{Challenges and Future Work.} Our current choice of autoencoders for generative modeling limits the ability to generate more fine-grained variations in environments. Using more sophisticated models based on GANs may improve the quality of generated environments. In addition, our current approach learns possible perturbations from the training dataset --- one potential direction is to augment this approach by prescribing a set of possible perturbations and training the generative model to select/combine the provided perturbations \cite{zakharov2019deceptionnet, robey2020model, wong2020learning}. 
% We are also interested in combining our approach with domain randomization techniques to generate larger-scale environment perturbations.

% \begin{table}[h]
% \caption{An Example of a Table}
% \label{table_example}
% \begin{center}
% \begin{tabular}{|c||c|}
% \hline
% One & Two\\
% \hline
% Three & Four\\
% \hline
% \end{tabular}
% \end{center}
% \end{table}

% \begin{figure}[thpb]
%   \centering
%   \framebox{\parbox{3in}{We suggest that you use a text box to insert a graphic (which is ideally a 300 dpi TIFF or EPS file, with all fonts embedded) because, in an document, this method is somewhat more stable than directly inserting a picture.
% }}
%   %\includegraphics[scale=1.0]{figurefile}
%   \caption{Inductance of oscillation winding on amorphous
%   magnetic core versus DC bias magnetic field}
%   \label{figurelabel}
% \end{figure}

%                                   % the textheight of the last page by a suitable amount.
%                                   % This command does not take effect until the next page
%                                   % so it should come on the page before the last. Make
%                                   % sure that you do not shorten the textheight too much.

%%%%%%%%%%%%%%%%%%%%%%%%%%%%%%%%%%%%%%%%%%%%%%%%%%%%%%%%%%%%%%%%%%%%%%%%%%%%%%%%

\section*{ACKNOWLEDGMENT}
The Toyota Research Institute (TRI) partially supported this work. This article solely reflects the opinions and conclusions of its authors and not TRI or any other Toyota entity. The authors were also partially supported by the Office of Naval Research [Award Number: N00014-18-1-2873], the NSF CAREER Award [2044149], and the School of Engineering and Applied Science at Princeton University through the generosity of William Addy ’82.
%%%%%%%%%%%%%%%%%%%%%%%%%%%%%%%%%%%%%%%%%%%%%%%%%%%%%%%%%%%%%%%%%%%%%%%%%%%%%%%%

% \begin{thebibliography}{99}

\putbib[bib-distributional-robustness]
\end{bibunit}

%==================================================
\renewcommand{\thetable}{A\arabic{table}}
\renewcommand{\theequation}{A\arabic{equation}}
\renewcommand\thefigure{A\arabic{figure}}
\renewcommand{\thesubsection}{A\arabic{subsection}}
\setcounter{figure}{0}
\setcounter{table}{0}
\setcounter{equation}{0}

\clearpage
\begin{bibunit}[corlabbrvnat]
\section*{Appendix}
\label{sec:appendix}

\subsection{Additional baseline and ablation studies}
\label{app:ablation}

\setlength{\tabcolsep}{3pt} % Make spacing between table columns smaller; default value: 6pt
\begin{table*}[h!]
\scriptsize
\vspace{-5pt}
\begin{center}
\begin{tabular}{cccccccc}
    \toprule
    & \multicolumn{6}{c}{Test - digit}\\
    \cmidrule{2-8}
    {Method} & 2 & 3 & 4 & 5 & 6 & 7 & 9\\ \midrule
    DRAGEN & \bftab 0.608 $\pm$ 0.016 & \bftab 0.787 $\pm$ 0.021 & \bftab 0.626 $\pm$ 0.023 & \bftab 0.589 $\pm$ 0.018 & \bftab 0.842 $\pm$ 0.012 & \bftab 0.841 $\pm$ 0.014 & \bftab 0.793 $\pm$ 0.018\\
    No norm loss & 0.532 $\pm$ 0.022 & 0.762 $\pm$ 0.028 & 0.604 $\pm$ 0.027 & 0.512 $\pm$ 0.019 & 0.751 $\pm$ 0.023 & 0.792 $\pm$ 0.021 & 0.735 $\pm$ 0.026 \\
    \bottomrule
\end{tabular}
\vspace{5pt}
\caption{Normalized reward (mean and standard deviation) for DRAGEN and DRAGEN without latent norm constraint using the pendulum task. Results for DRAGEN (top row) are the same as in Table.~\ref{tab:pendulum}}
\vspace{-15pt}
\label{tab:ablation}
\end{center}
\end{table*}

\textbf{Effect of constraining magnitudes of latent variables.} We remove the norm loss $\mathcal{L}_\text{norm}$ from the loss function \eqref{eq:embedding-loss} when training the embedding in the pendulum task. The norm loss is necessary because if the latent variables can grow in magnitude without any regularization, the Lipschitz continuity condition can be trivially satisfied, and then the Wasserstein distance around existing environments in the latent space will not capture shifts in task-relevant features. The results are shown in Table.~\ref{tab:ablation} below. Without the regularization, the baseline performs worse than DRAGEN among all digits.

\textbf{Effect of target ascent in cost when generating adversarial environments.} As described in Sec. \ref{sec:optimization}, when performing gradient ascent on the latent space, we set the target ascent (threshold for the increase in predicted cost) using a percentage $\Delta \tilde{C}_p \in [0,1]$ of $\mathcal{R}(\tilde{C})$, the empirical range of cost of all environments. Here we experiment different values of $\Delta \tilde{C}_p$ in the pendulum task with digit 6. The results are shown in Table.~\ref{tab:ablation-target-increase} below. When target ascent is small, perturbations of the environments are too small and OoD generalization is limited; when target ascent is too big, DRAGEN training becomes highly unstable as generated environments are too difficult, and OoD performance deteriorates significantly. 
\begin{table}[h!]
\scriptsize
\begin{center}
\begin{tabular}{cc}
    $\Delta \tilde{C}_p$ & Reward\\
    \midrule
    0.05 & 0.785 $\pm$ 0.015 \\
    0.1 & 0.783 $\pm$ 0.014 \\
    0.2 & \bftab 0.842 $\pm$ 0.012 \\
    0.4 & 0.654 $\pm$ 0.025 \\
    0.5 & 0.514 $\pm$ 0.033 \\
    \bottomrule
\end{tabular}
\vspace{5pt}
\caption{Normalized reward (mean and standard deviation) with different levels of target ascent using digit 3 with colored background in the pendulum task.}
\vspace{-10pt}
\label{tab:ablation-target-increase}
\end{center}
\end{table}

\textbf{Training using digit images with colored background.} In Sec.~\ref{sec:experiments} we hypothesize that DRAGEN will outperform Perlin noise in larger margins if colored distractions are added to the black background. Here we add uniform, randomly sampled color to the background of both MNIST (training) and USPS (testing) digit images. We perform experiments using digit 3, which is the only digit where DRAGEN performs worse than Perlin noise in regular USPS images with black background (Table \ref{tab:pendulum}).  DRAGEN achieves $0.715  \pm  0.023\%$ in normalized reward while Perlin noise only achieves $0.627 \pm  0.025\%$. We suspect that with black background, the structured Perlin noise "augments" the white digit (Fig.~\ref{fig:digit-comparison}), which is less effective if the background is colored.

\subsection{Experiment details}
\label{app:experiment}

\subsubsection{Pendulum task}
\textbf{Control policy training.} We use an open-sourced implementation of the Soft Actor Critic (SAC) algorithm (\url{https://github.com/astooke/rlpyt}). Some tricks \cite{kostrikov2020image} are also found to be helpful: (1) actor and critic share the same convolutional layers, and only the critic updates them; (2) before training starts, a fixed number of steps (10$\%$ of the replay buffer size) are collected with the initial policy, and these steps are kept in the replay buffer throughout training (to alleviate catastrophic forgetting). During re-training at each iteration, actor and critic models are saved at the step when the reward is the highest, and then loaded at the next iteration. Replay buffer is re-used between iterations.

For both settings (Landmark and Digit), the same policy network structure is used: two convolutional layers with 16, 32 channels, 6, 4 kernel sizes, 4, 2 strides, zero paddings; additionally the actor has two linear layers each with 128 units, and the critic has two each with 256 units; ReLU activation is applied at all layers. Other hyperparameters are listed below.

\textbf{Cost of an image for DRAGEN training.} For normalizing the cost of an environment between $[0,1]$, the lower bound corresponds to the pendulum not moving at all with itself hanging downwards, and the upper bound corresponds to the cost when the policy is trained using the true states of the pendulum instead of the camera image.

% critic target update frequency: 

\begin{table}[h!]
\footnotesize
\captionsetup{font=footnotesize}
\centering
\begin{tabular}{cc}
    \toprule
    Parameter & Value \\ \midrule
    Stacked frames & 3 \\
    Optimizer & Adam \\
    Learning rate & 1e-3 \\
    Discount factor & 0.99 \\
    Temperature (fixed) & 0.2 \\
    Gradient clipping norm & 3000 \\
    Num steps (per re-training) & 20000 \\
    Batch size & 256 \\
    Replay buffer size & 50000 \\
    Replay ratio & 32 \\
    \bottomrule
\end{tabular}
\vspace{5pt}
\caption{Hyperparameters of policy training for the pendulum task.}
\label{tab:pendulum-sac-hyperparameters}
\end{table}
% \vspace{-10pt}

\textbf{Generative modeling.} The encoder and decoder of the autoencoder follow a symmetric pattern: the same number $L$ of layers are used; the encoder uses convolutional layers with 4 kernel size, 2 stride, and 1 padding, which downsizes the image by a factor of 2 at each layer; the decoder uses convolutional layers with 3 kernel size, 1 stride, and 1 padding, which preserves the image size, but uses upsampling with a scale factor of 2 after each convolutional layer. We find that a good autoencoder with low reconstruction loss is essential to generating images of high qualities, and thus we run more epochs of embedding training at the first iteration of DRAGEN training. For all baselines and DRAGEN, the same number $K$ of new images is generated at each iteration. Additional hyperparameters are shown below.

% inner channels, predictor width,  num epocg before first gen, num epochg per gen,
% \vspace{-20pt}
\begin{table}[h!]
\footnotesize
\captionsetup{font=footnotesize}
\centering
\begin{tabular}{ccc}
    \toprule
    & \multicolumn{2}{c}{Setting} \\
    \cmidrule{2-3}
    Parameter & Landmark & Digit \\ \midrule
    Batch size & 4 & 8 \\
    Image size & 96 & 48 \\
    Observation size & 64 & 32 \\
    Latent dimension & 16 & 16 \\
    $\Delta \tilde{C}_p$ & 0.20 & 0.20 \\
    $L$ & \multicolumn{2}{c}{3} \\
    $K$ & \multicolumn{2}{c}{64} \\
    $\alpha_1$ & \multicolumn{2}{c}{0.1} \\
    $\alpha_2$ & \multicolumn{2}{c}{1.0} \\
    $\alpha_3$ & \multicolumn{2}{c}{0.1} \\
    $\overline{\gamma}$ & \multicolumn{2}{c}{0.04} \\
    Optimizer & \multicolumn{2}{c}{Adam} \\
    Learning rate & \multicolumn{2}{c}{1e-3} \\
    \bottomrule
\end{tabular}
\vspace{5pt}
\caption{Hyperparameters of DRAGEN training for the pendulum task.}
\label{tab:pendulum-dragen-hyperparameters}
\end{table}

% \vspace{-20pt}
\textbf{Sample training and testing images from datasets}

\begin{figure}[H]
\centering
\includegraphics[width=0.4\textwidth]{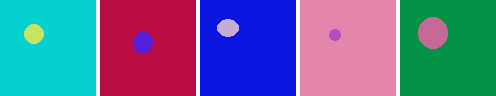}
\caption{Samples of landmark images - Normal, Closer, Farther, Smaller, Larger (from left to right)}
\label{fig:app-landmark-samples}
\end{figure}

\begin{figure}[H]
\centering
\includegraphics[width=0.4\textwidth]{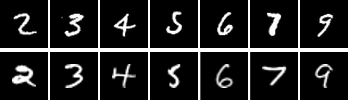}
\caption{Samples of digit images. Top: MNIST dataset (training) \cite{lecun1998gradient}; bottom: USPS dataset (testing) \cite{hull1994database}. The USPS dataset contains human drawings with more cursive digits.}
\label{fig:app-digit-samples}
\end{figure}

\begin{table}[h!]
\footnotesize
\captionsetup{font=footnotesize}
\centering
\begin{tabular}{cc}
    \toprule
    Parameter & Value \\ \midrule
    Optimizer & Adam \\
    Learning rate & 3e-3 \\
    % Weight decay & 1e-5 \\
    Num steps (per re-training) & 5000 \\
    Batch size & 128 \\
    Replay buffer size & 1000 \\
    Replay ratio & 4 \\
    \bottomrule
\end{tabular}
\vspace{5pt}
\caption{Hyperparameters of policy training for the grasping task.}
\label{tab:grasping-policy-hyperparameters}
\end{table}

\subsubsection{Grasping in simulation}
\textbf{Control policy training.} The fully convolutional network (FCN) that acts as the Q function first consists of three convolutional layers that each downsamples the image by a factor of 2, and then three convolutional layers with bilinear upsampling to restore the original dimensions of the input image. A 1x1 convolutional layer is applied at the end to set RGB channels of the output. ReLU activation and batch normalization are applied at all layers except for the last one. Since grasp is executed only at the chosen pixel of the output prediction maps, gradients of binary cross-entropy loss are only passed through that pixel. During re-training at each iteration, we apply $\epsilon$-greedy exploration with $\epsilon$ initialized at 1.0 (all actions are random initially) and annealed to 0.2. Policy model is saved at the step when the training reward is the highest, and then loaded at the next iteration. Replay buffer is re-used between iterations. Other hyperparameters are listed below.

\vspace{-10pt}
\begin{figure}[H]
\centering
\includegraphics[width=0.2\textwidth]{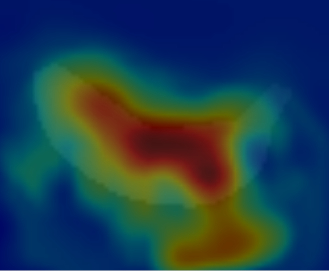}
\caption{Sample pixel-wise prediction of grasp success of a banana-like object.}
\label{fig:app-affordance}
\end{figure}

\textbf{Generative modeling.} The PointNet \cite{qi2017pointnet} encoder consists of three 1D convolutional layers, spatial softmax, and three linear layers to encode an object from its sampled surface points to a low-dimensional latent variable. The decoder network consists of six linear layers and a skip connection at the fourth layer and it decodes the SDF value from a pair of the query 3D location and latent variable. For each training object (from datasets or generated ones), 2048 surface points and SDF values at 20000 locations are sampled (implementation of SDF sampling is from \url{https://github.com/marian42/mesh_to_sdf}). For each batch of training data, a fixed number of objects are sampled along with all 2048 surface points and 2048 SDF values out of the 2000 available ones for each object. Same as in the pendulum task, more epochs of embedding training are run at the first iteration to train a good autoencoder with low reconstruction loss. For Domain Randomization and DRAGEN, the same number $K$ of new objects is generated at each iteration. Additional hyperparameters are shown below.

\begin{table}[h!]
\footnotesize
\captionsetup{font=footnotesize}
\centering
\begin{tabular}{cc}
    \toprule
    Parameter & Value \\ \midrule
    Latent dimension & 128 \\
    $K$& 128 \\
    Batch size & 4 \\
    $\Delta \tilde{C}_p$ & 0.1 \\
    $\alpha_1$ & 0.1 \\
    $\alpha_2$ & 1.0 \\
    $\alpha_3$ & 0.1 \\
    $\overline{\gamma}$ & 0.04 \\
    Optimizer & Adam \\
    Learning rate & 1e-3 \\
    \bottomrule
\end{tabular}
\vspace{5pt}
\caption{Hyperparameters of DRAGEN training for the grasping task.}
\label{tab:grasping-dragen-hyperparameters}
\end{table}

\textbf{Cost of an object in DRAGEN training.} We initially trained the cost predictor using binary grasping cost (successfully lifted or not) but gradient ascent on the latent space did not generate diverse, task-relevant variations in object shapes. Thus we assign a more continuous cost to an object based on the minimum friction coefficient needed for a successful grasp - each object is evaluated at 10 different friction coefficient values between 0.10 and 0.55 with 0.05 interval (stop if success), and the 10 values corresponds to cost values from 0.0 to 0.9 with 0.1 interval. If a grasp fails with the highest friction coefficient (0.55), the cost is 1.0. Thus there are a total of 11 possible cost values for an object. We hypothesize that using higher resolution of cost values can further improve DRAGEN's performance, but it comes at the cost of running more simulation at each iteration when environments are re-evaluated before the generative model and embedding are re-trained.

\subsubsection{Grasping on hardware}
Experiments are performed using a Franka Panda arm and a Microsoft Azure Kinect RGB-D camera. Robot Operating System (ROS) Melodic package (on Ubuntu 18.04) is used to integrate robot arm control and perception.

% \begin{figure}[H]
% \centering
% \includegraphics[width=0.5\textwidth]{figures-appendix/2d_objects_new.jpg}
% \caption{10 toy bricks used in hardware experiments for 2D grasping. Objects that were failed to be grasped by policies trained using (1) DRAGEN: 9; (2) Domain Randomization (DR): 6, 7, 9; (3) No data augmentation: 0, 6, 7, 9. Object 9 is very different from training objects (triangles, ellipses, and rectangles). Generally triangles are more difficult to grasp than ellipses and rectangles. DRAGEN policies were able to grasp object 6, a relatively large triangle that requires precise grasp at its largest corner, while the other policies failed to grasp it (see video for the trials).}
% \label{fig:app-2d-objects}
% \end{figure}

\begin{figure}[H]
\centering
\includegraphics[width=0.48\textwidth]{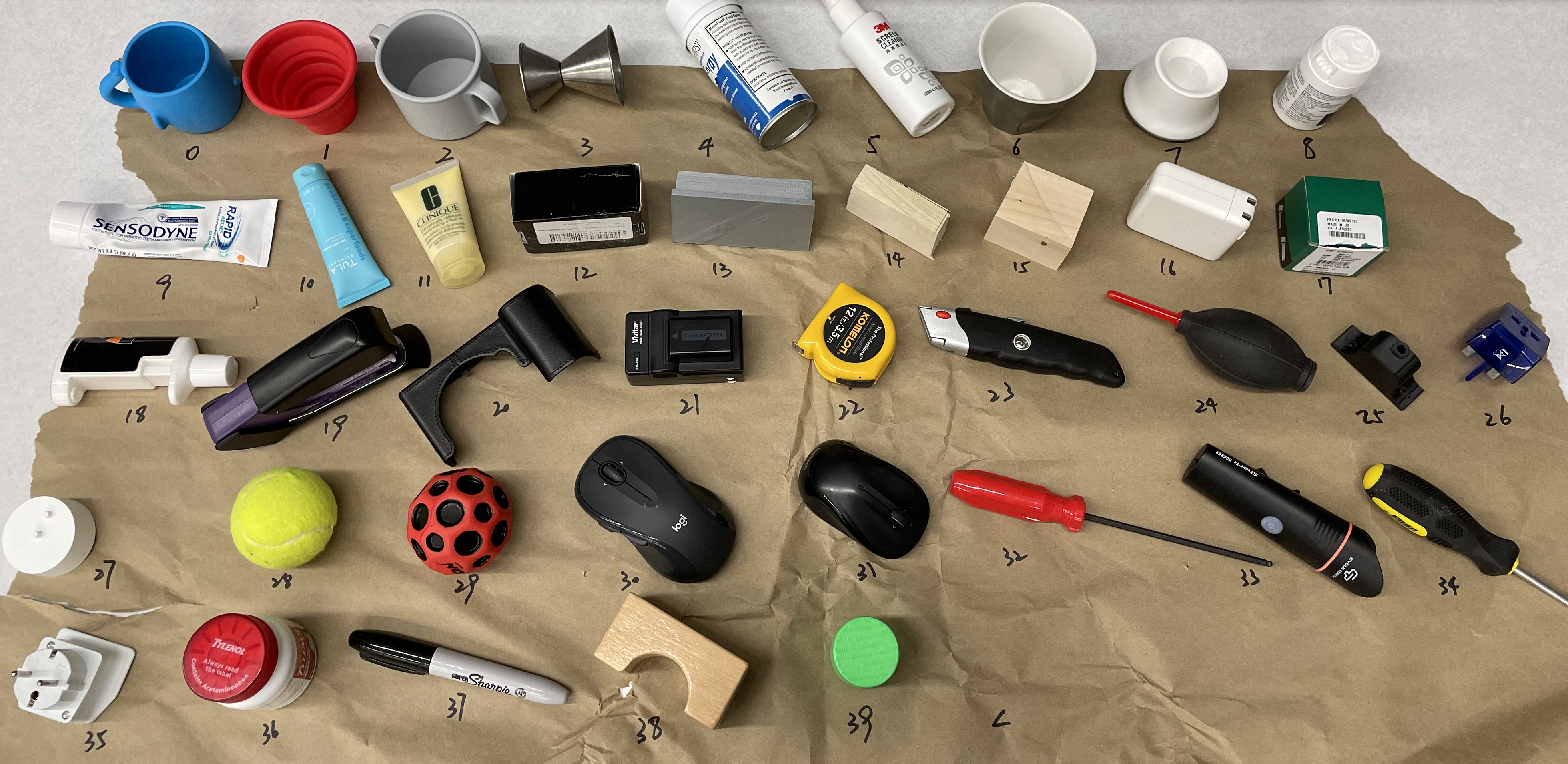}
\caption{40 common objects used in hardware experiments for 3D grasping.}
\label{fig:app-3d-objects}
\end{figure}

\begin{figure}[H]
\centering
\includegraphics[width=0.48\textwidth]{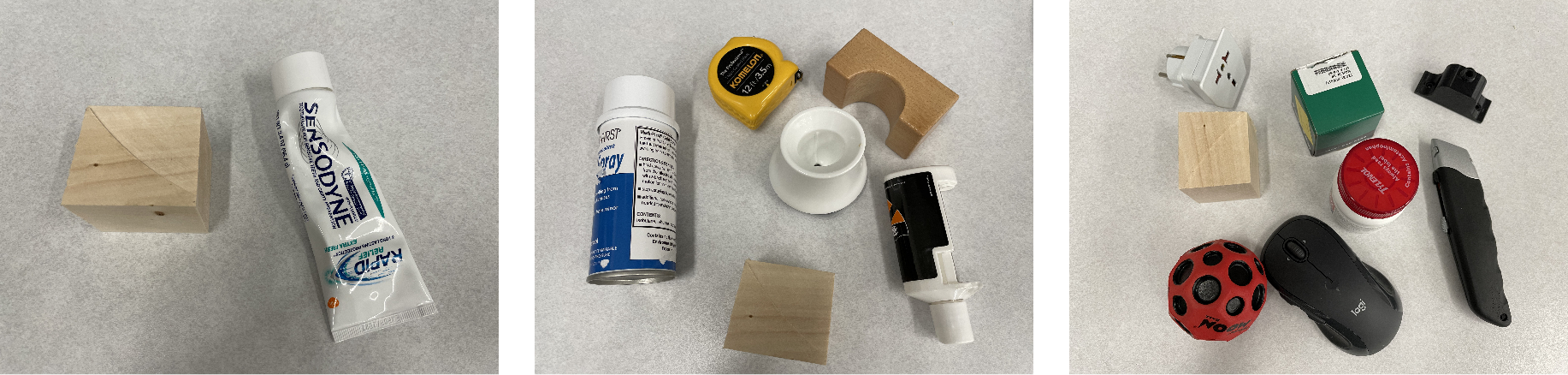}
\caption{Objects that were failed to be grasped by policies trained using (left) DRAGEN, (middle) Domain Randomization, (right) no data augmentation. The wooden cube causes failure for all policies as both its length and width are very close to the gripper width.}
\label{fig:app-3d-objects-failed}
\end{figure}

\putbib[bib-distributional-robustness]
\end{bibunit}

% \end{thebibliography}

\begin{thebibliography}{10}
\providecommand{\url}[1]{#1}
\csname url@rmstyle\endcsname
\providecommand{\newblock}{\relax}
\providecommand{\bibinfo}[2]{#2}
\providecommand\BIBentrySTDinterwordspacing{\spaceskip=0pt\relax}
\providecommand\BIBentryALTinterwordstretchfactor{4}
\providecommand\BIBentryALTinterwordspacing{\spaceskip=\fontdimen2\font plus
\BIBentryALTinterwordstretchfactor\fontdimen3\font minus
  \fontdimen4\font\relax}
\providecommand\BIBforeignlanguage[2]{{%
\expandafter\ifx\csname l@#1\endcsname\relax
\typeout{** WARNING: IEEEtran.bst: No hyphenation pattern has been}%
\typeout{** loaded for the language `#1'. Using the pattern for}%
\typeout{** the default language instead.}%
\else
\language=\csname l@#1\endcsname
\fi
#2}}

\bibitem{sunderhauf2018limits}
N.~S{\"u}nderhauf, O.~Brock, W.~Scheirer, R.~Hadsell, D.~Fox, J.~Leitner,
  B.~Upcroft, P.~Abbeel, W.~Burgard, M.~Milford, \emph{et~al.}, ``The limits
  and potentials of deep learning for robotics,'' \emph{The International
  Journal of Robotics Research}, vol.~37, no. 4-5, pp. 405--420, 2018.

\bibitem{sinha2017certifying}
A.~Sinha, H.~Namkoong, and J.~Duchi, ``Certifiable distributional robustness
  with principled adversarial training,'' in \emph{Proceedings of the
  International Conference on Learning Representations (ICLR)}, 2018.

\bibitem{blanchet2019quantifying}
J.~Blanchet and K.~Murthy, ``Quantifying distributional model risk via optimal
  transport,'' \emph{Mathematics of Operations Research}, vol.~44, no.~2, pp.
  565--600, 2019.

\bibitem{esfahani2018data}
P.~M. Esfahani and D.~Kuhn, ``Data-driven distributionally robust optimization
  using the {W}asserstein metric: Performance guarantees and tractable
  reformulations,'' \emph{Mathematical Programming}, vol. 171, no.~1, pp.
  115--166, 2018.

\bibitem{namkoong2016stochastic}
H.~Namkoong and J.~C. Duchi, ``Stochastic gradient methods for distributionally
  robust optimization with $f$-divergences.'' in \emph{Proceedings of the
  Advances in Neural Information Processing Systems (NIPS)}, vol.~29, 2016, pp.
  2208--2216.

\bibitem{volpi2018generalizing}
R.~Volpi, H.~Namkoong, O.~Sener, J.~C. Duchi, V.~Murino, and S.~Savarese,
  ``Generalizing to unseen domains via adversarial data augmentation,'' in
  \emph{Proceedings of the Advances in Neural Information Processing Systems
  (NIPS)}, 2018, pp. 5334--5344.

\bibitem{tobin2017domain}
J.~Tobin, R.~Fong, A.~Ray, J.~Schneider, W.~Zaremba, and P.~Abbeel, ``Domain
  randomization for transferring deep neural networks from simulation to the
  real world,'' in \emph{Proceedings of the IEEE/RSJ International Conference
  on Intelligent Robots and Systems (IROS)}, 2017, pp. 23--30.

\bibitem{mehta2020active}
B.~Mehta, M.~Diaz, F.~Golemo, C.~J. Pal, and L.~Paull, ``Active domain
  randomization,'' in \emph{Proceedings of the Conference on Robot Learning
  (CoRL)}, 2020, pp. 1162--1176.

\bibitem{tobin2018domain}
J.~Tobin, L.~Biewald, R.~Duan, M.~Andrychowicz, A.~Handa, V.~Kumar, B.~McGrew,
  A.~Ray, J.~Schneider, P.~Welinder, \emph{et~al.}, ``Domain randomization and
  generative models for robotic grasping,'' in \emph{Proceedings of the
  IEEE/RSJ International Conference on Intelligent Robots and Systems (IROS)},
  2018, pp. 3482--3489.

\bibitem{kostrikov2020image}
D.~Yarats, I.~Kostrikov, and R.~Fergus, ``Image augmentation is all you need:
  Regularizing deep reinforcement learning from pixels,'' in \emph{Proceedings
  of the International Conference on Learning Representations (ICLR)}, 2021.

\bibitem{laskin2020reinforcement}
M.~Laskin, K.~Lee, A.~Stooke, L.~Pinto, P.~Abbeel, and A.~Srinivas,
  ``Reinforcement learning with augmented data,'' in \emph{Proceedings of the
  Advances in Neural Information Processing Systems (NeurIPS)}, vol.~33, 2020,
  pp. 19\,884--19\,895.

\bibitem{dennis2020emergent}
M.~Dennis, N.~Jaques, E.~Vinitsky, A.~Bayen, S.~Russell, A.~Critch, and
  S.~Levine, ``Emergent complexity and zero-shot transfer via unsupervised
  environment design,'' in \emph{Proceedings of the Advances in Neural
  Information Processing Systems (NeurIPS)}, vol.~33, 2020, pp.
  13\,049--13\,061.

\bibitem{wang2019paired}
R.~Wang, J.~Lehman, J.~Clune, and K.~O. Stanley, ``Paired open-ended
  trailblazer ({POET}): Endlessly generating increasingly complex and diverse
  learning environments and their solutions,'' \emph{arXiv preprint
  arXiv:1901.01753}, 2019.

\bibitem{goodfellow2014explaining}
I.~Goodfellow, J.~Shlens, and C.~Szegedy, ``Explaining and harnessing
  adversarial examples,'' in \emph{Proceedings of the International Conference
  on Learning Representations (ICLR)}, 2015.

\bibitem{jalal2017robust}
A.~Jalal, A.~Ilyas, C.~Daskalakis, and A.~G. Dimakis, ``The robust manifold
  defense: Adversarial training using generative models,'' \emph{arXiv preprint
  arXiv:1712.09196}, 2017.

\bibitem{robey2020model}
A.~Robey, H.~Hassani, and G.~J. Pappas, ``Model-based robust deep learning,''
  \emph{arXiv preprint arXiv:2005.10247}, 2020.

\bibitem{wong2020learning}
E.~Wong and J.~Z. Kolter, ``Learning perturbation sets for robust machine
  learning,'' in \emph{Proceedings of the International Conference on Learning
  Representations (ICLR)}, 2021.

\bibitem{zakharov2019deceptionnet}
S.~Zakharov, W.~Kehl, and S.~Ilic, ``Deception{N}et: Network-driven domain
  randomization,'' in \emph{Proceedings of the IEEE/CVF International
  Conference on Computer Vision (ICCV)}, 2019, pp. 532--541.

\bibitem{lee2020shapeadv}
K.~Lee, Z.~Chen, X.~Yan, R.~Urtasun, and E.~Yumer, ``Shape{A}dv: Generating
  shape-aware adversarial 3{D} point clouds,'' \emph{arXiv preprint
  arXiv:2005.11626}, 2020.

\bibitem{wang2019adversarial}
D.~Wang, D.~Tseng, P.~Li, Y.~Jiang, M.~Guo, M.~Danielczuk, J.~Mahler,
  J.~Ichnowski, and K.~Goldberg, ``Adversarial grasp objects,'' in
  \emph{Proceedings of the IEEE International Conference on Automation Science
  and Engineering (CASE)}, 2019, pp. 241--248.

\bibitem{morrison2020egad}
D.~Morrison, P.~Corke, and J.~Leitner, ``{EGAD}! an evolved grasping analysis
  dataset for diversity and reproducibility in robotic manipulation,''
  \emph{IEEE Robotics and Automation Letters}, vol.~5, no.~3, pp. 4368--4375,
  2020.

\bibitem{joshi2019semantic}
A.~Joshi, A.~Mukherjee, S.~Sarkar, and C.~Hegde, ``Semantic adversarial
  attacks: Parametric transformations that fool deep classifiers,'' in
  \emph{Proceedings of the IEEE/CVF International Conference on Computer Vision
  (ICCV)}, 2019, pp. 4773--4783.

\bibitem{park2019deepsdf}
J.~J. Park, P.~Florence, J.~Straub, R.~Newcombe, and S.~Lovegrove, ``Deep{SDF}:
  Learning continuous signed distance functions for shape representation,'' in
  \emph{Proceedings of the IEEE Conference on Computer Vision and Pattern
  Recognition (CVPR)}, 2019, pp. 165--174.

\bibitem{achlioptas2018learning}
P.~Achlioptas, O.~Diamanti, I.~Mitliagkas, and L.~Guibas, ``Learning
  representations and generative models for 3{D} point clouds,'' in
  \emph{Proceedings of the International Conference on Machine Learning
  (ICML)}, 2018, pp. 40--49.

\bibitem{hinton2006reducing}
G.~E. Hinton and R.~R. Salakhutdinov, ``Reducing the dimensionality of data
  with neural networks,'' \emph{Science}, vol. 313, no. 5786, pp. 504--507,
  2006.

\bibitem{villani2009optimal}
C.~Villani, \emph{Optimal transport: old and new}.\hskip 1em plus 0.5em minus
  0.4em\relax Springer, 2009, vol. 338.

\bibitem{arjovsky2017wasserstein}
M.~Arjovsky, S.~Chintala, and L.~Bottou, ``Wasserstein generative adversarial
  networks,'' in \emph{Proceedings of the International Conference on Machine
  Learning (ICML)}, 2017, pp. 214--223.

\bibitem{ren2021distributionally}
A.~Z. Ren and A.~Majumdar, ``Distributionally robust policy learning via
  adversarial environment generation,'' \emph{arXiv preprint arXiv:2107.06353},
  2021.

\bibitem{lecun1998gradient}
Y.~LeCun, L.~Bottou, Y.~Bengio, and P.~Haffner, ``Gradient-based learning
  applied to document recognition,'' \emph{Proceedings of the IEEE}, vol.~86,
  no.~11, pp. 2278--2324, 1998.

\bibitem{hull1994database}
J.~J. Hull, ``A database for handwritten text recognition research,''
  \emph{IEEE Transactions on Pattern Analysis and Machine intelligence},
  vol.~16, no.~5, pp. 550--554, 1994.

\bibitem{haarnoja2018soft}
T.~Haarnoja, A.~Zhou, P.~Abbeel, and S.~Levine, ``Soft {A}ctor-{C}ritic:
  Off-policy maximum entropy deep reinforcement learning with a stochastic
  actor,'' in \emph{Proceedings of the International Conference on Machine
  Learning (ICML)}, 2018, pp. 1861--1870.

\bibitem{perlin1985image}
K.~Perlin, ``An image synthesizer,'' \emph{ACM Siggraph Computer Graphics},
  vol.~19, no.~3, pp. 287--296, 1985.

\bibitem{coumans2021}
E.~Coumans and Y.~Bai, ``Py{B}ullet, a {P}ython module for physics simulation
  for games, robotics and machine learning,'' \url{http://pybullet.org},
  2016--2021.

\bibitem{wohlkinger20123dnet}
W.~Wohlkinger, A.~Aldoma, R.~B. Rusu, and M.~Vincze, ``3{DN}et: Large-scale
  object class recognition from {CAD} models,'' in \emph{Proceedings of the
  IEEE International Conference on Robotics and Automation (ICRA)}, 2012, pp.
  5384--5391.

\bibitem{zeng2018learning}
A.~Zeng, S.~Song, S.~Welker, J.~Lee, A.~Rodriguez, and T.~Funkhouser,
  ``Learning synergies between pushing and grasping with self-supervised deep
  reinforcement learning,'' in \emph{Proceedings of the IEEE/RSJ International
  Conference on Intelligent Robots and Systems (IROS)}, 2018, pp. 4238--4245.

\bibitem{qi2017pointnet}
C.~R. Qi, H.~Su, K.~Mo, and L.~J. Guibas, ``Point{N}et: Deep learning on point
  sets for 3{D} classification and segmentation,'' in \emph{Proceedings of the
  IEEE Conference on Computer Vision and Pattern Recognition (CVPR)}, 2017, pp.
  652--660.

\bibitem{kleineberg2020adversarial}
M.~Kleineberg, M.~Fey, and F.~Weichert, ``Adversarial generation of continuous
  implicit shape representations,'' \emph{arXiv preprint arXiv:2002.00349},
  2020.

\end{thebibliography}
\end{document}